\documentclass[conference,a4paper]{IEEEtran}

\usepackage{silence}
\IEEEoverridecommandlockouts

\usepackage[hidelinks]{hyperref}
\usepackage[cmex10]{amsmath}
\usepackage{amssymb,amsfonts}
\usepackage{graphicx}
\graphicspath{{figures/}}

\usepackage{booktabs}
\usepackage{siunitx}
\usepackage[numbers,compress]{natbib}
\usepackage{texnames}
\usepackage{bm,bbm}
\usepackage{orcidlink}

\usepackage{subcaption}
\usepackage{hyperref,natbib}
\usepackage{amsmath}
\usepackage{amsthm}
\usepackage{thmtools,thm-restate}
\usepackage{bigints}
\usepackage{amssymb,amsopn,algorithm,algorithmic,float,bbm,bm,enumerate,color,multirow}
\usepackage{makecell}

\definecolor{nodata}{RGB}{0,0,0}
\definecolor{maize}{RGB}{255,211,0}
\definecolor{soybean}{RGB}{37,111,0}
\definecolor{rice}{RGB}{0,168,226}
\definecolor{wheat}{RGB}{137,96,83}
\definecolor{other}{RGB}{128,128,128}

\begin{document}

\title{\uppercase{On the Generalizability of Foundation Models for Crop Type Mapping}
\thanks{Copyright 2024 IEEE. Published in the 2025 IEEE International Geoscience and Remote Sensing Symposium (IGARSS 2025), scheduled for 3 - 8 August 2025 in Brisbane, Australia. Personal use of this material is permitted. However, permission to reprint/republish this material for advertising or promotional purposes or for creating new collective works for resale or redistribution to servers or lists, or to reuse any copyrighted component of this work in other works, must be obtained from the IEEE. Contact: Manager, Copyrights and Permissions / IEEE Service Center / 445 Hoes Lane / P.O. Box 1331 / Piscataway, NJ 08855-1331, USA. Telephone: + Intl. 908-562-3966.\vspace{0.5em}}
\thanks{The work was supported in part by the National Science Foundation (NSF) through awards IIS 21-31335, OAC 21-30835, DBI 20-21898, C3.ai research award, and Taiwan-UIUC Fellowship. This work made use of the Illinois Campus Cluster, a computing resource that is operated by the Illinois Campus Cluster Program (ICCP) in conjunction with the National Center for Supercomputing Applications (NCSA) and which is supported by funds from the University of Illinois at Urbana-Champaign.}
}

\author{
Yi-Chia Chang\textsuperscript{1}\orcidlink{0000-0002-8881-1996}, 
Adam J. Stewart\textsuperscript{2,3}\orcidlink{0000-0002-0468-5006}, 
Favyen Bastani\textsuperscript{4}\orcidlink{0000-0002-1100-4192}, 
Piper Wolters\textsuperscript{4}\orcidlink{0009-0007-0837-2801}, \\
Shreya Kannan\textsuperscript{1},
George R. Huber\textsuperscript{1},
Jingtong Wang\textsuperscript{1},
Arindam Banerjee\textsuperscript{1}\orcidlink{0009-0003-8211-6989}
\\
\textsuperscript{1}\textit{University of Illinois Urbana-Champaign}, 
\textsuperscript{2}\textit{Technical University of Munich}, \\
\textsuperscript{3}\textit{Munich Center for Machine Learning},
\textsuperscript{4}\textit{Allen Institute for AI} \\
\{yichia3, shreya28, ghuber3, jw132, arindamb\}@illinois.edu, \\
adam.stewart@tum.de, \{favyenb, piperw\}@allenai.org
}

\maketitle

\begin{abstract}
Foundation models pre-trained using self-supervised learning have shown powerful transfer learning capabilities on various downstream tasks, including language understanding, text generation, and image recognition. The Earth observation (EO) field has produced several foundation models pre-trained directly on multispectral satellite imagery for applications like precision agriculture, wildfire and drought monitoring, and natural disaster response. However, few studies have investigated the ability of these models to generalize to new geographic locations, and potential concerns of geospatial bias---models trained on data-rich developed nations not transferring well to data-scarce developing nations---remain. We evaluate three popular EO foundation models, SSL4EO-S12, SatlasPretrain, and ImageNet, on five crop classification datasets across five continents. Results show that pre-trained weights designed explicitly for Sentinel-2, such as SSL4EO-S12, outperform general pre-trained weights like ImageNet. While only 100 labeled images are sufficient for achieving high overall accuracy, 900 images are required to mitigate class imbalance and improve average accuracy. 
\end{abstract}


\section{Introduction}

Crop type maps have many critical downstream applications in food security~\citep{foodsecurity} and conservation, including crop yield prediction~\citep{khaki2019crop,doraiswamy2003crop,van2020crop,Fan2022cropyield}, understanding interactions between wildlife and crop fields~\citep{gross2018seasonality,pywell2015wildlife}, and regional crop damage assessment~\citep{silleos2002assessment,rahman2020systematic}. However, the availability and accuracy of crop type maps varies greatly across regions: the US and EU maintain large-scale, regularly updated maps with 80+\% accuracy~\citep{eucropmapaccuracy,luman2008assessment}, but crop type maps of most other regions are sporadically updated or incomplete due to weaker crop type self-reporting policies, less government resources, and a greater prevalence of smallholder farming that is more difficult to map. Even sparse high-accuracy crop type labels are rare in these regions. Thus, while numerous methods have been proposed to map crop types through remote sensing~\citep{prins2021crop,adrian2021sentinel,tang2024deep,yuan2023bridging}, these have seldom been deployed globally, since high-quality training data does not exist for most of the world. 

Transfer learning has been applied to remote sensing for agricultural monitoring and land cover mapping~\citep{tlma, generalizedfewshot}, including detecting paddy rice in Spain and France and summer barley in the Netherlands~\citep{TL2024}, agricultural field boundary delineation~\citep{fow}, and geographic generalizability in tree detection~\citep{sachdeva2024distribution}. Pre-trained foundation models have shown promising generalizability on downstream tasks with remote sensing data~\citep{presto, cropharvest, fewshotglobal}. Remote sensing foundation models like SSL4EO-S12~\citep{ssl4eos12} and SatlasPretrain~\citep{satlaspretrain} pre-trained on globally available data, such as Sentinel-1/2, present a key opportunity for increasing the accuracy of global crop type mapping by improving the ability for models trained on downstream tasks to generalize across regions. However, transfer learning for crop type mapping in concert with such foundation models has not been studied at scale, in part because of the lack of a harmonized dataset of global crop type labels.

\begin{table*}[htbp]
    \centering
    \caption{A list of the datasets used in this study. All datasets are harmonized and subsampled to 1,000 patches for fair comparison. Number of classes includes ``other'' but excludes ``nodata''. Resolution is measured in meters per pixel. Labels collected by ground survey or self-declaration are considered to be close to 100\% accurate, while ML-labeled datasets are listed with their reported overall accuracy on the principal crop classes. We calculate the harmonized dataset accuracy based on the average of all datasets. Datasets are listed alphabetically.}
    \label{tab:datasets}
    \begin{tabular}{lcrccr}
        \toprule
        Dataset & Region & \# Classes & Resolution (m) & Accuracy & License \\
        \midrule
        Cropland Data Layer (CDL)~\citep{cdl} & USA & 134 & 30 & 84.1\% & CC0-1.0 \\
        EuroCrops~\citep{eurocrops} & Europe & 331 & - & Self-declared & CC-BY-SA-4.0 \\
        Northeastern China Crop Map (NCCM)~\citep{nccm} & China & 4 & 10 & 87.0\% & CC-BY-4.0 \\
        South Africa Crop Type (SACT)~\citep{sact} & South Africa & 9 & 10 & Ground survey & CC-BY-4.0 \\
        South America Soybean (SAS)~\citep{sas} & South America & 2 & 30 & 95.3\% & Unknown \\ \midrule
        Harmonized & Global & 5 & 10 & 93.3\% & CC-BY-SA-4.0 \\
        \bottomrule
    \end{tabular}
\end{table*}

In this paper, we first present a harmonized global crop type mapping dataset that incorporates five regional datasets across five continents. We identify the available data sources with the highest label quality in each continent and pair their crop type labels with a uniform collection of cloud-free Sentinel-2 images captured during the peak of the growing season in each region. We focus on four major cereal grains and harmonize the class labels of each dataset around these categories. We then experiment with three popular pre-trained models and find that SSL4EO-S12 provides the best results. Further experiments with varying amounts of in-distribution and out-of-distribution data highlight the need for additional datasets in data-scarce regions like Africa and South America. All experiments are implemented using TorchGeo~\citep{torchgeo}, all preprocessed datasets and code are available for download from Hugging Face\footnote{\url{https://huggingface.co/datasets/torchgeo/harmonized_global_crops}} and GitHub\footnote{\url{https://github.com/yichiac/crop-type-transfer-learning}}.

\section{Data Curation}

Dozens of existing crop type classification datasets exist in the literature, with varying label quality and crop type categories. While a few small-scale datasets are manually labeled by experts, the majority of large-scale datasets are generated by machine learning models and must first be vetted by experts before they can reliably be used for further model training. To facilitate the training of global crop type classification models, we manually inspected and verified the validity of all publicly available crop type classification datasets and selected five datasets from five continents noteworthy for their size and label accuracy. Table~\ref{tab:datasets} lists all five datasets, as well as statistics about the dataset size and estimated accuracy levels published by their original authors.

Since each dataset includes a different number of crop classes with differing levels of granularity, we first harmonize all mask labels to six classes: 0: no data (unknown), 1: maize (corn), 2: soybean, 3: rice, 4: wheat, and 5: other (known). Maize, soybean, rice, and wheat are chosen as the four major cereal grains in our study. These four crop types represent the four most valuable crops, making up the vast majority of all global cereal production and roughly half of agricultural lands worldwide~\cite{martin2019regional}. They are also prevalent in all of the datasets we chose, albeit with very different frequencies. We also reserve classes for ``other'', including both other forms of agriculture and non-agricultural land use, and ``nodata'', where the land cover type is unknown and may or may not include one of our four crop types of interest. All nodata pixels are ignored in this study when computing metrics so as to avoid unfairly penalizing the model. 

After class harmonization, we chip each dataset into \(256 \times 256\)~px patches. As the majority of these datasets are published as raster and vector mask layers without any corresponding imagery, we download our own Sentinel-2 imagery for each labeled region from Google Cloud\footnote{\url{https://cloud.google.com/storage/docs/public-datasets/sentinel-2}}. For each mask, we download a cloud-free Sentinel-2 L1C image patch for the same location during the peak of the growing season (depending on latitude) during the year the mask was acquired. This is also done for the few datasets that do come with imagery so as to ensure that all Sentinel-2 spectral bands are present in the image. All images and masks are then warped to a Web Mercator projection at 10~m/px resolution. Although time series information would likely improve model performance, we do not consider it in this study as not all models support satellite image time series. For computational feasibility and better representing the real-world limited data availability, we subsampled 1,000 images for each region, resulting in 5,000 images in total.

\section{Methodology}

\subsection{ResNet-50 with U-Net}
A ResNet-50 backbone with a U-Net architecture is proposed to extract the spatial features of satellite imagery for crop type mapping. ResNet~\citep{resnet}, consisting of convolutional layers and residual blocks, is a popular deep neural network for feature extraction in computer vision tasks. We chose ResNet-50 because of its increased model complexity, ability to capture representations, and available pre-trained weights. Using ResNet-50 as the encoder for extracting features can enhance the generalizability and training efficiency in downstream domains, especially with limited labeled data. The decoder of U-Net~\citep{unet} operates on top of the feature maps extracted by the ResNet-50 backbone. During all experiments, we freeze the ResNet-50 encoder weights and fine-tune the U-Net decoder weights, allowing for faster training and aligning with our transfer learning goals.

\subsection{Pre-trained weights}

The following pre-trained weights are evaluated in this study:

\subsubsection{SSL4EO-S12\texorpdfstring{~\citep{ssl4eos12}}{[SSL4EO-S12]}}
provides ResNet and Vision Transformer (ViT)~\citep{vit} backbones pre-trained using MoCo-v2~\citep{mocov2} and DINO~\citep{dino}. The global pre-training dataset includes Sentinel-1 GRD and Sentinel-2 L1C/L2A imagery.

{SatlasPretrain~\citep{satlaspretrain}} 
\subsubsection{SatlasPretrain\texorpdfstring{~\citep{satlaspretrain}}{[SatlasPretrain]}}
provides ResNet and Swin Transformer~\citep{swintransformer} backbones pre-trained on global Landsat 8/9, Sentinel-1/2, and National Agriculture Imagery Program (NAIP) images. 

\subsubsection{ImageNet\texorpdfstring{~\citep{imagenet}}{[ImageNet]}}
serves as a baseline due to its accessibility and wide use in computer vision research. This baseline allows us to compare performance against models that were not trained on satellite imagery.

\subsection{Transfer learning}

We perform several transfer learning experiments, including in-domain and out-of-domain evaluation. Here, we define in-domain (ID) as data from the same geographic region the model was trained on, and out-of-distribution (OOD) as data from different geographic regions not seen during training. In the ID experiments, we seek to understand how performance improves with increased ID data, ranging from 10 to 900 ID samples. In the OOD experiments, we investigate the effects of combining OOD data with varying amounts of ID data. In these experiments, we use all OOD data, meaning up to 4,000 samples from other regions. We change the number of ID training samples to analyze their contribution to model performance when OOD is available. A 90--10 train--test split is used for all experiments.

\section{Results and discussion}

\begin{table}[htbp]
\centering
\caption{Average F1 scores for different pre-trained weights and ID sample sizes. SSL4EO-S12 consistently outperforms the other two pre-trained weights in all regions. Increasing ID data improves model performance.}
\label{table:id_evaluation}
\begin{tabular}{lcccccc}
\toprule
Weights & ImageNet & SatlasPretrain & \multicolumn{3}{c}{SSL4EO-S12} \\
\midrule
\# Samples & 900 ID & 900 ID & 10 ID & 100 ID & 900 ID \\
\midrule
CDL       & 0.58 & 0.57 & 0.28 & 0.38 & \textbf{0.77} \\
EuroCrops & 0.39 & 0.35 & 0.26 & 0.30 & \textbf{0.48} \\
NCCM      & 0.59 & 0.37 & 0.17 & 0.38 & \textbf{0.74} \\
SACT      & 0.32 & 0.43 & 0.18 & 0.16 & \textbf{0.70} \\
SAS       & 0.82 & 0.74 & 0.22 & 0.27 & \textbf{0.85} \\
\bottomrule
\end{tabular}
\end{table}

\subsection{Pre-trained weights comparison}

In the first experiment, we explore various pre-trained weights to find one that performs well for crop type mapping under the ID setting. Table~\ref{table:id_evaluation} shows that SSL4EO-S12 consistently outperforms all other weights by 3--27\% in all study regions. In contrast, SatlasPretrain demonstrates suboptimal performance compared to other models. This is likely because this model only accepts 9 out of 13 spectral bands as input, whereas SSL4EO-S12 accepts all 13, and ImageNet weights can easily be repeated using libraries like timm~\citep{rw2019timm}. All models show lower performance on EuroCrops and SACT, where no-data pixels are most abundant. However, SSL4EO-S12 demonstrates robustness in these data-scarce regions.

\begin{figure}[htbp]
    \centering
    \includegraphics[width=\columnwidth]{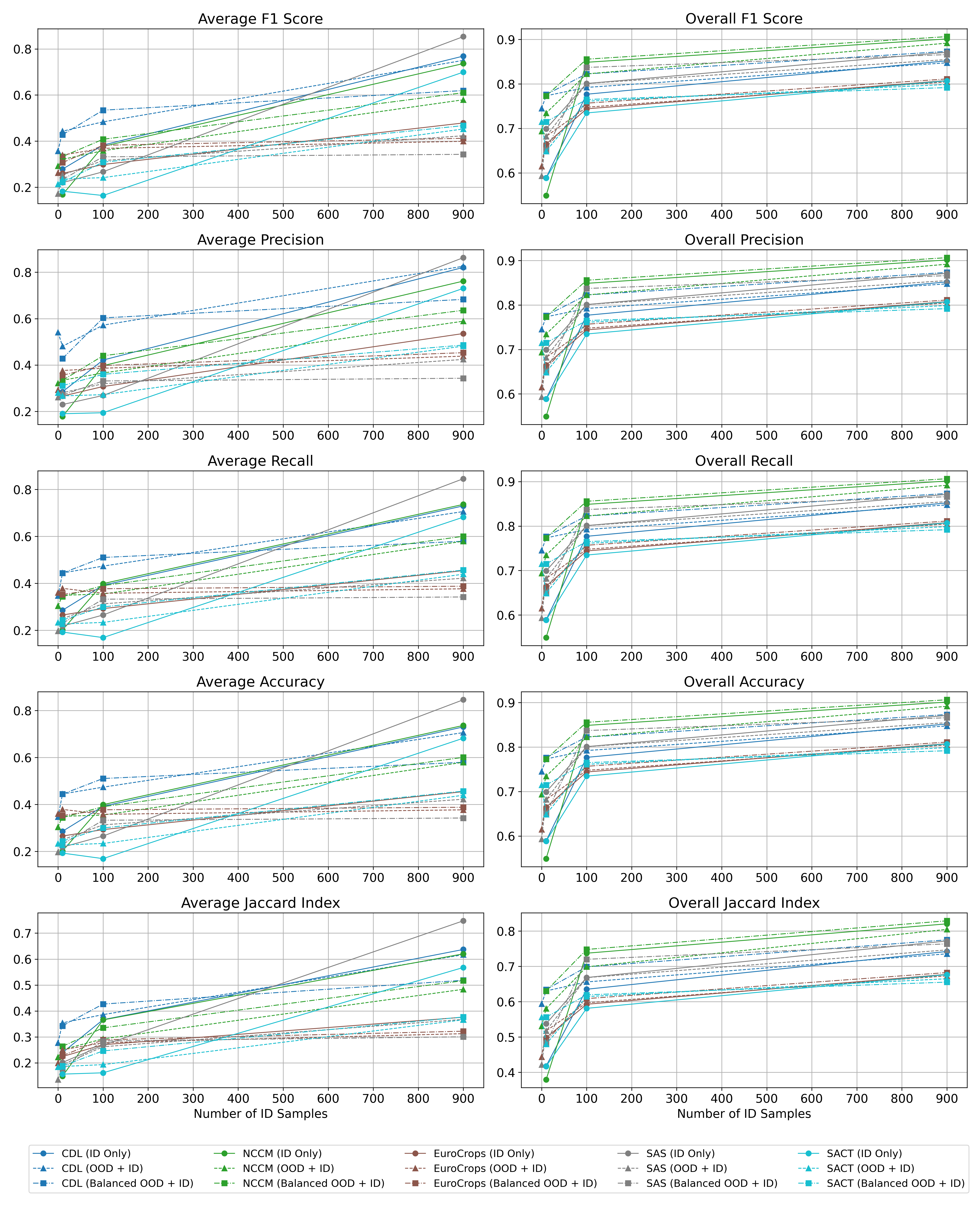}
    \caption{Reported metrics of ID, OOD + ID, and balanced OOD + ID using SSL4EO-S12 pre-trained weights. Average and overall metrics are given for F1-score, precision, recall, accuracy, and Jaccard Index (IoU). For all metrics, higher is better.}
    \label{fig:metrics}
\end{figure}

\subsection{In-domain evaluation} 
We use the best-performing pre-trained weights (SSL4EO-S12) to explore the effects of ID training data size on model performance. Table~\ref{table:id_evaluation} and Figure~\ref{fig:metrics} show that all regions benefit from increasing ID data, as expected. However, while 100 samples is sufficient to achieve high \textit{overall} accuracy, 900 samples are required to achieve high \textit{average} accuracy. The additional samples appear to be necessary for the model to overcome the extreme class imbalance present in all datasets.

\begin{table*}[htbp]
\centering
\caption{Average F1 scores for ID--OOD samples using SSL4EO-S12 pre-trained weights. The results show the benefits of pre-training OOD with limited ID data and the importance of prioritizing ID samples when more ID samples are available.}
\label{table:OOD}
\begin{tabular}{lccccccc}
\toprule
 & 4000 OOD & \multicolumn{3}{c}{4000 OOD} &3990 OOD & 3900 OOD & 3100 OOD \\
\cmidrule(lr){3-5} \cmidrule(lr){6-8}
Dataset &  & +10 ID & +100 ID & +900 ID & +10 ID & +100 ID & +900 ID \\
\midrule
CDL       & 0.36 & 0.44 & 0.48 & \textbf{0.75} & 0.43 & 0.53 & \textbf{0.62} \\
EuroCrops & 0.26 & 0.34 & 0.37 & \textbf{0.40} & 0.31 & 0.38 & \textbf{0.41} \\
NCCM      & 0.29 & 0.32 & 0.36 & \textbf{0.58} & 0.33 & 0.41 & \textbf{0.61} \\
SACT      & 0.21 & 0.24 & 0.24 & \textbf{0.45} & 0.22 & 0.31 & \textbf{0.47} \\
SAS       & 0.17 & 0.25 & 0.32 & \textbf{0.42} & 0.23 & 0.33 & \textbf{0.34} \\
\bottomrule
\end{tabular}
\end{table*}

\subsection{Few-shot learning evaluation}

We perform three experiments on OOD data: zero shot learning where only OOD data is used for training, few shot learning where OOD data is combined with increasing amounts of ID data, and a balanced variation of few shot learning where the same total amount of training data is used for all ID/OOD splits. The results of these experiments are presented in Table~\ref{table:OOD}.

The advantages of pre-training on OOD data are particularly profound in scenarios with limited ID samples. By comparing Tables~\ref{table:id_evaluation} and \ref{table:OOD}, we see that models trained with OOD samples without any ID samples (zero shot) can already outperform models trained with 10 ID samples. This shows the benefits of pre-trained weights with OOD data, which can alleviate the difficulties when ID data is unavailable in target regions. 

However, the results also show that OOD can cause worse performance when more ID samples become available. For example, a model trained on 4,000 OOD + 900 ID samples performs worse than a model trained solely on 900 ID samples. This is due to the distribution shift between ID and OOD data, as OOD data tends to have different class balance and field shapes compared to the target region. We recommend using OOD data to pre-train the model, but fine-tuning solely on ID data to handle this distribution shift.

While the most common scenario in remote sensing is a large fixed-size OOD dataset and varying amounts of ID data, it can be difficult to separate accuracy improvements due to additional ID data from improvements due to additional training data (OOD + ID). We repeat the above experiment with a fixed training size and varying ratios of ID and OOD data. While there are slight variations in the average F1 scores, the general trend remains the same.

\subsection{Qualitative evaluation}
\begin{figure}[htbp]
    \centering
    \renewcommand{\arraystretch}{1}
    \vbox{
        \begin{tabular}{lll}
            \colorbox{nodata}{\rule{0pt}{3pt}\rule{3pt}{0pt}} \enskip No Data &
            \colorbox{maize}{\rule{0pt}{3pt}\rule{3pt}{0pt}} \enskip Maize &
            \colorbox{soybean}{\rule{0pt}{3pt}\rule{3pt}{0pt}} \enskip Soybean \\
            \colorbox{rice}{\rule{0pt}{3pt}\rule{3pt}{0pt}} \enskip Rice &
            \colorbox{wheat}{\rule{0pt}{3pt}\rule{3pt}{0pt}} \enskip Wheat &
            \colorbox{other}{\rule{0pt}{3pt}\rule{3pt}{0pt}} \enskip Other
        \end{tabular}
    }
    \vspace{2pt}
    \setlength\tabcolsep{1.5pt}
    \renewcommand{\arraystretch}{1}
    \begin{tabular}{cccc}
        \rotatebox[origin=l,y=8mm]{90}{CDL} &
        \includegraphics[width=0.3\linewidth]{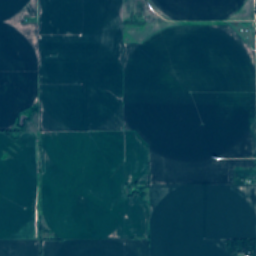} &
        \includegraphics[width=0.3\linewidth]{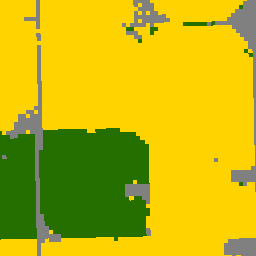} &
        \includegraphics[width=0.3\linewidth]{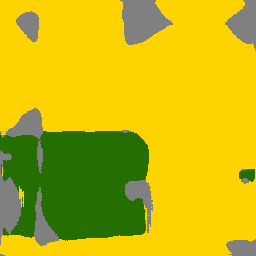} \\
        \rotatebox[origin=l,y=4mm]{90}{EuroCrops} &
        \includegraphics[width=0.3\linewidth]{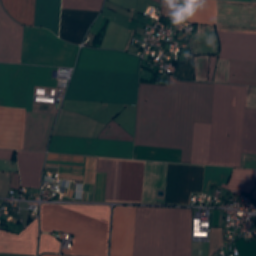} &
        \includegraphics[width=0.3\linewidth]{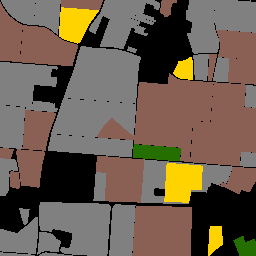} &
        \includegraphics[width=0.3\linewidth]{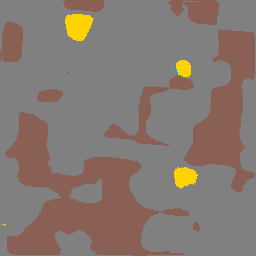} \\
        \rotatebox[origin=l,y=7mm]{90}{NCCM} &
        \includegraphics[width=0.3\linewidth]{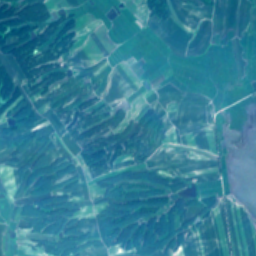} &
        \includegraphics[width=0.3\linewidth]{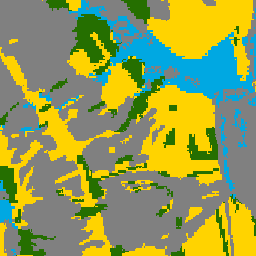} &
        \includegraphics[width=0.3\linewidth]{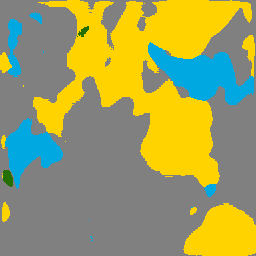} \\
        
        \rotatebox[origin=l,y=8mm]{90}{SACT} &
        \includegraphics[width=0.3\linewidth]{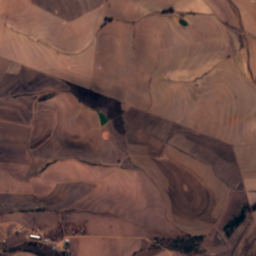} &
        \includegraphics[width=0.3\linewidth]{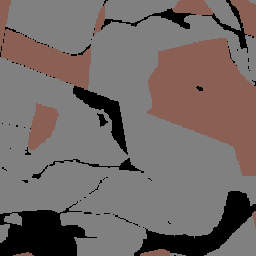} &
        \includegraphics[width=0.3\linewidth]{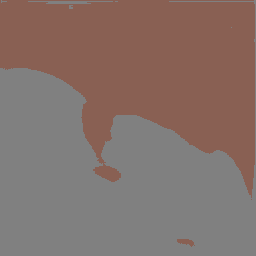} \\
        \rotatebox[origin=l,y=9mm]{90}{SAS} &
        \includegraphics[width=0.3\linewidth]{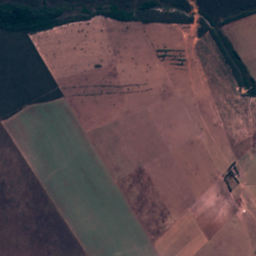} &
        \includegraphics[width=0.3\linewidth]{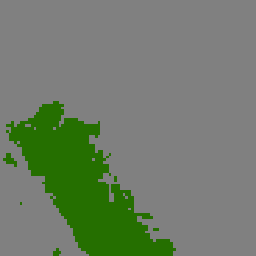} &
        \includegraphics[width=0.3\linewidth]{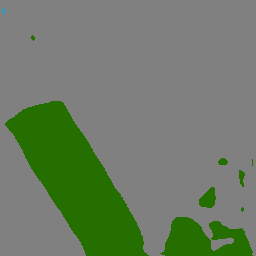} \\
        & Image & Mask & Prediction
    \end{tabular}
    \caption{Visualization of example input Sentinel-2 images, ground truth masks, and model predictions using SSL4EO-S12 pre-trained weights. Overall results are promising, with the models capturing the general class distribution and correctly identifying most fields.
}
    \label{fig:prediction}
\end{figure}

Figure~\ref{fig:prediction} shows example predictions for each dataset. The most obvious visual difference between mask and prediction is the loss of fine-scale features like roads, as seen in prior works on CDL crop classification using foundation models~\cite{stewart2024ssl4eo}. This is especially evident for NCCM, where intercropping (the practice of planting rows of different crops in close proximity) is seen for maize and soybeans. In the EuroCrops and SACT predictions, we see that the model has difficulty distinguishing wheat from ``other'', possibly due to the similar resemblance of weeds.

\section{Conclusion}

In this work, we present a global semantic segmentation dataset for crop type mapping created by harmonizing five crop classification datasets from five continents. We investigate the ability of various pre-trained foundation models to perform crop classification in ID and ID + OOD sample settings. We find that models like SSL4EO-S12 that are pre-trained on all spectral bands of Sentinel-2 imagery outperform other competing models. We also find that OOD data can improve performance in regions with limited ID data. However, training on both ID and OOD data can actually hurt model performance if care is not taken.

Larger datasets are still needed, especially for regions like Africa and South America, where large multi-class datasets are scarce. However, dataset size is not the only important factor, with class imbalance plaguing all datasets used in this study. To correctly predict uncommon classes with higher precision, especially when models move from the four most important cereal grains to hundreds of agricultural classes, the issue of class imbalance will need to be addressed, either through weighted dataset sampling or weighted loss functions.

\small
\bibliographystyle{IEEEtranN}
\bibliography{references.bib}

\end{document}